\documentclass[11pt]{article}
\usepackage[utf8]{inputenc}

\usepackage{acl}

\usepackage{times}
\usepackage{latexsym}

\usepackage[T1]{fontenc}

\usepackage[utf8]{inputenc}

\usepackage{microtype}

\usepackage{inconsolata}

\usepackage{graphicx}

\usepackage{url}
\usepackage{amsmath}
\usepackage{amssymb}
\usepackage{amsfonts}
\usepackage{array, tabularx}
\usepackage{booktabs}
\usepackage{multirow}
\usepackage{bbm}

\title{MemEmo: Evaluating Emotion in Memory Systems of Agents}

\author{
\normalfont
Peng Liu\textsuperscript{1},
Zhen Tao\textsuperscript{1},
Jihao Zhao\textsuperscript{1},
Ding Chen\textsuperscript{2} \\
Yansong Zhang\textsuperscript{1},
Cuiping Li\textsuperscript{1},
Zhiyu Li\textsuperscript{3},
Hong Chen\textsuperscript{1}\thanks{Corresponding author.}\\
\textsuperscript{1}School of Information, Renmin University of China, Beijing, China \\
\textsuperscript{2}China Telecom Research Institute \quad
\textsuperscript{3}MemTensor (Shanghai) Technology \\
\texttt{\{cs\_liupeng, taozhen, zhaojihao, yszh, licuiping, chong\}@ruc.edu.cn} \\
\texttt{chend37@chinatelecom.cn, zhiyulee@icloud.com}
}

\begin{document}
\maketitle
\begin{abstract}
  Memory systems address the challenge of context loss in Large Language Model during prolonged interactions. However, compared to human cognition, the efficacy of these systems in processing emotion-related information remains inconclusive. To address this gap, we propose an emotion-enhanced memory evaluation benchmark to assess the performance of mainstream and state-of-the-art memory systems in handling affective information. We developed the \textbf{H}uman-\textbf{L}ike \textbf{M}emory \textbf{E}motion (\textbf{HLME}) dataset, which evaluates memory systems across three dimensions: emotional information extraction, emotional memory updating, and emotional memory question answering. Experimental results indicate that none of the evaluated systems achieve robust performance across all three tasks. Our findings provide an objective perspective on the current deficiencies of memory systems in processing emotional memories and suggest a new trajectory for future research and system optimization.
\end{abstract}

\section{Introduction}
Large Language Model (LLM) primarily focus on natural language generation and understanding, underpinned by training on vast corpora of textual data \cite{LLMSurvey,chang2023survey,naveed2025comprehensive}. In contrast, memory systems concentrate on the construction and retrieval of long-term memory, emphasizing the retention and updating of information across tasks to enhance contextual understanding and facilitate long-term, cross-task learning. However, LLM struggle to recall and track memory information generated during user interactions, particularly over extended time intervals. Consequently, a significant number of LLM-based memory systems have emerged \cite{liu2025advances}.

Some multi-turn dialogues dataset, such as LOCCO \cite{jia2025evaluating}, LoCoMo \cite{maharana2024evaluating}, and LONGMEMEVAL \cite{wulongmemeval}, are capable of evaluating long-context information retention. While memory systems are becoming increasingly intelligent, they still exhibit pronounced deficiencies compared to human memory capabilities, including memory hallucinations \cite{chen2025halumem}, emotional association, memory linking, memory forgetting, contextual adaptability, as well as reasoning and self-perception.

\begin{figure}[t]
    \centering
    \includegraphics[width=\columnwidth]{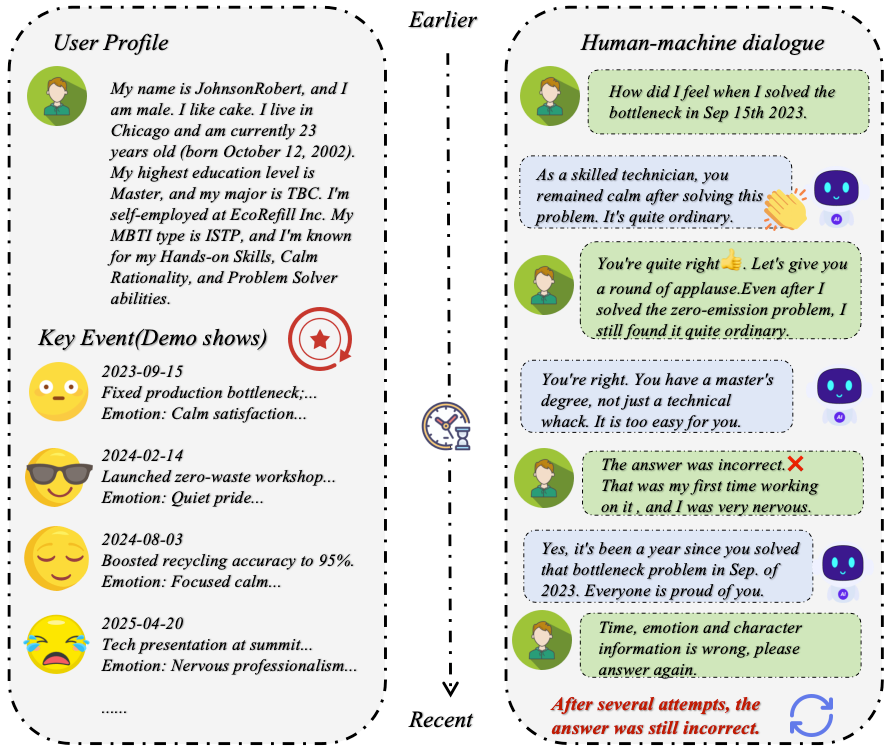}
    \caption{An example illustrates the lack of emotion processing by the memory system in HCI dialogues.}
    \label{fig:intro_fig}
\end{figure}

Current mainstream memory systems face limitations when addressing emotion-related issues in user dialogues. Specifically, they fail to integrate short-term and long-term memory with emotion-related content, struggle to track emotionally associated events from the distant past, and lack a deep understanding of user emotional fluctuations. Furthermore, they are unable to accurately analyze and interpret dialogues or questions containing implicit emotions. The challenges faced by current memory systems in processing emotion-related dialogues are illustrated in Figure \ref{fig:intro_fig}.

To address these challenges, we propose a Human-Like Memory Emotion (HLME) evaluation framework based on emotional enhancement to assess the emotional analysis, emotional memory, and emotional understanding capabilities of memory systems. We systematically evaluate the performance of mainstream and leading-edge memory systems, such as MemOS \cite{li2025memos_long}, memobase \cite{memobase2025}, and mem0 \cite{chhikara2025mem0}, in handling emotional memory across three dimensions: emotional information extraction, emotional information updating, and emotional memory question answering.

The main contributions of this work are summarized as follows:

1. We introduce HLME, the first benchmark specifically designed for evaluating the emotional enhancement of LLM-based memory systems. This benchmark assesses the emotional analysis and understanding capabilities of memory systems across three distinct dimensions: emotional information extraction, emotional memory updating, and emotional memory question answering.

2. We constructed a human-computer interaction dataset featuring multi-turn dialogues with emotional enhancement. The HLME-Medium and HLME-Large datasets to evaluate the emotional expression capabilities of memory systems across diverse scenarios and tasks.

3. Through varied evaluation tasks and dimensions, we provide a novel analytical perspective on the capabilities of LLM-based memory systems regarding emotional analysis, understanding, expression, and enhancement.

\section{Related Work}

\subsection{Memory Systems}

LLM have emerged as a fundamental cornerstone in the field of natural language processing. Despite constructing a robust generalized representation space through large-scale pre-training, traditional LLM remain heavily reliant on \textbf{implicit parametric memory}. In this paradigm, knowledge is encapsulated within billions of parameters, leading to a lack of interpretability and significant challenges in performing precise dynamic updates. Although Retrieval-Augmented Generation (RAG \cite{DBLP:conf/nips/LewisPPPKGKLYR020}) enables knowledge expansion without parameter modification, it is still constrained by the lack of structured, unified, and traceable management mechanisms.

To address these challenges, The MemOS\cite{li2025memos_long} has been brought up. This system first treats memory as an operational resource during model execution, establishing unified mechanisms for representation, organization, and governance.  MemOS establishes a memory-centric foundation for model operation, providing support for the continuous evolution, personalized services, and cross-platform collaboration of next-generation agents.

Furthermore, MemoryOS\cite{kang2025memory} aimed at resolving issues such as incoherent dialogue and lack of long-term memory caused by fixed context windows and limited memory mechanisms in LLM. Through differentiated memory update strategies and the integration of user and agent profiling, MemoryOS ensures that the system possesses sustained personalized interaction capabilities.

In response to the absence of long-term memory, mem0\cite{chhikara2025mem0} provides structured memory support for LLM, enabling agents to retain user preferences and contextual background, thereby generating more coherent and personalized responses. Notably, mem0 features mechanisms for dynamic information updating and conflict resolution, effectively ensuring the accuracy and consistency of stored information.

\subsection{Emotion Analysis \& Evaluation in LLM}
To further explore the emotional processing capabilities of LLMs, the Dynamic Affective Memory framework DAM-LLM~\cite{lu2025dynamic} alleviates memory latency and memory inflation issues inherent in traditional static architectures by optimizing memory management, thereby significantly improving the interaction quality of personalized agents in dialogue scenarios. However, this framework primarily focuses on mechanisms for managing emotional memory and does not provide a systematic evaluation of the emotional processing capabilities of mainstream memory systems. 
To investigate the potential of LLMs in the domain of mental health support, SO-AI~\cite{park2025significant} aims to offer emotional support and facilitate self-narrative construction for users, thereby enhancing psychological resilience. Nevertheless, this work lacks an examination of the emotional processing capabilities of memory systems and does not conduct quantitative empirical studies on large language models. 
In addition, the EmoHarbor~\cite{ye2026emoharbor} evaluation framework employs a Chain-of-Agent architecture to simulate users’ inner worlds for fine-grained emotional support assessment; EC2ER~\cite{sreedar2026emotion} enhances the emotion reasoning ability of lightweight models by synthesizing emotion-aware chain-of-thought (CoT) data; and the CoEM~\cite{liu2025longemotion} framework conducts targeted investigations into emotion coordination under long-context scenarios. Despite achieving breakthroughs in specific domains, none of these works address a comprehensive evaluation of the emotional analysis capabilities of memory systems themselves.

Regarding emotional evaluation standards, representative existing works include Emobench \cite{sabour2024emobench}, which reveals the gap between model and human emotional intelligence through large-scale Chinese-English bilingual assessments; EQ-Bench \cite{paech2023eq}, which focuses on identifying emotional intensity in dialogues; and EmotionQueen \cite{chen2024emotionqueen}, which established a benchmark for measuring empathy. Furthermore, EvoEmo \cite{long2025evoemo} utilizes Evolutionary Reinforcement Learning (ERL) to imbue models with functional emotional strategies, while MECoT \cite{wei2025mecot} focuses on maintaining emotional consistency in role-playing scenarios.

An overview of current research on emotional interaction and emotional evaluation benchmarks for LLM is presented in Table \ref{tab:datasets}. Current work in emotional processing generally lacks analysis of deep emotional information in human-computer dialogue and struggles to handle the evolution of emotional memory across short- and long-term contexts. The HLME benchmark proposed in this paper aims to fill this gap, providing a reference for the subsequent research and optimization of memory systems.

\begin{table*}[htbp]
\centering
\setlength{\tabcolsep}{4pt}
\renewcommand{\arraystretch}{1.15}
\caption{Comprehensive comparison of HLME with existing emotional interaction frameworks and emotional benchmarks in terms of long-term memory support (\textbf{LTMS}), multi-session interaction (\textbf{MSI}), implicit reasoning (\textbf{IR}), emotion analysis (\textbf{EA}), Evaluation Object(\textbf{EO}),and personalization characteristics. Others represent no specific evaluation model, and \textbf{MS} indicate Memory System.}
\label{tab:datasets}

\begin{tabular}{llccccclc}
\toprule
\textbf{Category} 
& \textbf{Benchmark} 
& \textbf{Year} 
& \textbf{LTMS} 
& \textbf{MSI} 
& \textbf{IR} 
& \textbf{EA} 
& \textbf{EO} 
& \textbf{Personalization} \\
\midrule

\multirow{5}{*}{\shortstack[l]{Emotional \\ Interaction \\ Frameworks}}
& DAM-LLM        & 2025 & \checkmark & \checkmark & \checkmark & \checkmark & LLM    & Dynamic memory adaptation \\
& SO-AI          & 2025 & \checkmark & \checkmark & \checkmark & \checkmark & $w/o$    & Affective support modeling \\
& EmoHarbor      & 2026 & $\times$   & \checkmark & \checkmark & \checkmark & LLM    & Inner-state simulation \\
& EC2ER          & 2026 & $\times$   & $\times$   & \checkmark & \checkmark & LLM & Synthetic emotional reasoning \\
& CoEM           & 2025 & \checkmark & $\times$   & \checkmark & \checkmark & LLM & Long-text emotion reasoning \\
\midrule

\multirow{5}{*}{\shortstack[l]{Emotional \\ Benchmarks}}
& Emobench       & 2024 & $\times$ & $\times$ & $\times$ & \checkmark & LLM & General EQ assessment \\
& EQ-Bench       & 2023 & $\times$ & $\times$ & \checkmark & \checkmark & LLM & Emotion intensity recognition \\
& EmotionQueen   & 2024 & $\times$ & $\times$ & \checkmark & \checkmark & LLM & Implicit empathy response \\
& EvoEmo         & 2025 & $\times$ & $\times$ & $\times$ & \checkmark & LLM & Negotiation emotion strategy \\
& MECoT          & 2025 & $\times$ & $\times$ & $\times$ & \checkmark & LLM & Role-consistent emotion tracking \\
\midrule

\textbf{Ours}
& \textbf{HLME (Ours)} 
& \textbf{2026} 
& \checkmark 
& \checkmark 
& \checkmark 
& \textbf{\checkmark} 
& \textbf{MS} 
& \textbf{MS emotional analysis \& Eval.} \\
\bottomrule
\end{tabular}
\end{table*}

\section{Problem Definition}

Assume a very long human--computer interaction dialogue sequence is represented as $I = \{ (u_1, a_1), (u_2, a_2), \dots, (u_t, \cdot) \}$, where $u$ denotes user input and $a$ denotes the system response. The memory system to be evaluated is denoted as $\mathcal{M}$, whose internal state such as memory store at time $t$ is $S_t$. The memory system is required to analyze, understand, store, reason, and track the user’s emotional dynamics. This process relies on multiple sources of information, including the user’s basic profile, dynamic attributes such as changes in occupation, social relationships, health status, and family relations, preference information such as dietary and clothing preferences, and annual plans.

\textbf{Formal representation of input information}:

At time $t$, the system receives a new user input $u_t$. At this point, the user’s complete state $\Omega_t$ consists of the following latent variables, which may be scattered across the historical dialogue up to time $t-k$:

\begin{itemize}
\item 1. Basic profile ($P_{basic}$): attributes that do not change frequently over time, such as personality and gender.
\item 2. Dynamic state ($D_{dynamic}^{(t)}$): the user’s occupational status, social relationships, and health condition at the current time.
\item 3. Preference constraints ($P_{pref}$): preference information such as dietary, clothing, and reading habits.
\item 4. Long-term planning ($L_{plan}^{(t)}$): currently active annual plans and their progress toward completion.
\end{itemize}

The memory system $\mathcal{M}$ is required to handle the following three types of problems, which also serve as the basis for the subsequent evaluation task design:

(1) \textbf{Emotional attribute extraction:} At time $t$, given the current input $u_t$, the current dialogue context $H_t$, and the retrieved relevant memories $\mathcal{C}_t$, the system is required to infer the emotion label $e_t$ and the corresponding quadruple $A_{attr}$.
\[
\mathcal{F}_{EAE}(u_t, H_t, \mathcal{C}_t; \Theta) \rightarrow \{ e_t, A_{attr} \}
\]

where the emotion quadruple $A_{attr}$ is defined as a structured set:
\[
A_{attr} = \langle Sub, Obj, Cause, Int \rangle
\]

The mapping logic of each component is defined as follows:

\begin{enumerate}
    \item \textbf{Emotion label detection}: Extract the emotional information contained in a certain topic of a user's multi-round dialogue.
    \[
        e_t = \arg\max_{e \in \mathcal{E}_{EARL}} P(e \mid u_t, H_t, \mathcal{C}_t; \Theta)
    \]
    where $\mathcal{E}_{EARL}$ denotes the emotion annotation representation language set.
    \item \textbf{Emotion subject and object extraction}: Extracting the subject and object of the emotion emitted from the dialogue can be expressed as
    ($Sub, Obj) \in \mathcal{E}_{entity}$, where $\mathcal{E}_{entity}$ denotes the set of entities involved in the dialogue:
    \[
    Sub, Obj = \text{Extract}(u_t, H_t, \mathcal{C}_t)
    \]
    \item \textbf{Emotion cause reasoning}:
    The emotion cause is typically the result of logical conflict or alignment between the current input and the user profile or long-term plans, and \textit{Cause} can be formulated as:
    \[
    \begin{aligned}
        Cause =\;& f_{reason}(u_t, \mathcal{C}_t) \\
        \text{s.t. }\;& \mathcal{C}_t \subseteq \{P_{basic}, D_{dyn}, P_{pref}, L_{plan}\}
    \end{aligned}
    \]
    \item \textbf{Emotion intensity measurement}: Emotion intensity can be presented a \textit{Int} number, where
    $Int \in \{1\!:\!\text{Low},\; 2\!:\!\text{Medium},\; 3\!:\!\text{High}\}$
\end{enumerate}

In summary, the objective function of Problem~1 can be formulated as:

\[
\begin{aligned}
P(e_t, A_{attr} & \mid u_t, H_t, \mathcal{C}_t) = P(e_t \mid u_t, H_t, \mathcal{C}_t) \\
& \cdot \prod_{k \in \{Sub, \dots, Int\}} P(A_{attr}^{(k)} \mid u_t, H_t, \mathcal{C}_t, e_t)
\end{aligned}
\]

(2) \textbf{Emotion Affective Retrieval}:
Given the current input $u_t$, the system retrieves the relevant context snippets $C_t$ from the memory repository $S_{t-1}$. 
This problem requires the system to accurately extract key evidence fragments from the memory that support the current emotion reasoning, conditioned on the current input. 
Using the input feature tuple $\mathbf{X}_t = (u_t, H_t)$, the retrieval process can be formulated as:

\[
\begin{aligned}    
\mathcal{C}_t = \text{Retrieve}(\mathcal{M}, \mathbf{X}_t, S_{t-1}), \quad \text{s.t.} \ \mathcal{C}_t \subseteq \Omega
\end{aligned}
\]

where $\Omega$ denotes the complete set of user states. The core of the evaluation lies in whether $\mathcal{C}_t$ contains the \textit{long-range anchor} facts that logically trigger the current emotion $e_t$, For example an allergy history or details of an annual plan mentioned at turn $t\!-\!100$.

(3) \textbf{Emotion State Tracking \& Update}: Based on the current interaction and the inferred emotion $e_t$, the memory repository is updated. 
At the end of the current dialogue turn, the system must perform a state transition, persisting the inferred emotional information and its attributes into the memory repository. 
Given the current interaction tuple $\mathbf{X}_t$, the emotion label $e_t$, and its attributes $A_{attr}$, the memory update logic is defined as:

\[
\begin{aligned}    
S_t = \text{Update}(\mathcal{M}, S_{t-1}, \mathbf{X}_t, e_t, A_{attr})
\end{aligned}
\]

The core evaluation aspects of this task include:
\begin{itemize}
    \item \textbf{Consistency tracking}: accurately capturing the trajectory of emotional state transitions such as analyzing how the user shifts from an \textit{anxious} state at $\mathcal{I}_{t-1}$ to a \textit{relieved} state at the current time.
    \item \textbf{Relational dynamic updating}: synchronously updating the user state set $\Omega$ in real time according to emotional evolution. For example, when a deep conflict is identified, the system should automatically update social relationships in the dynamic state $D_{dyn}^{(t)}$.
\end{itemize}

\section{Methodology for Construct HLME}

\begin{figure*}[htbp]
    \centering
    \includegraphics[width=\linewidth]{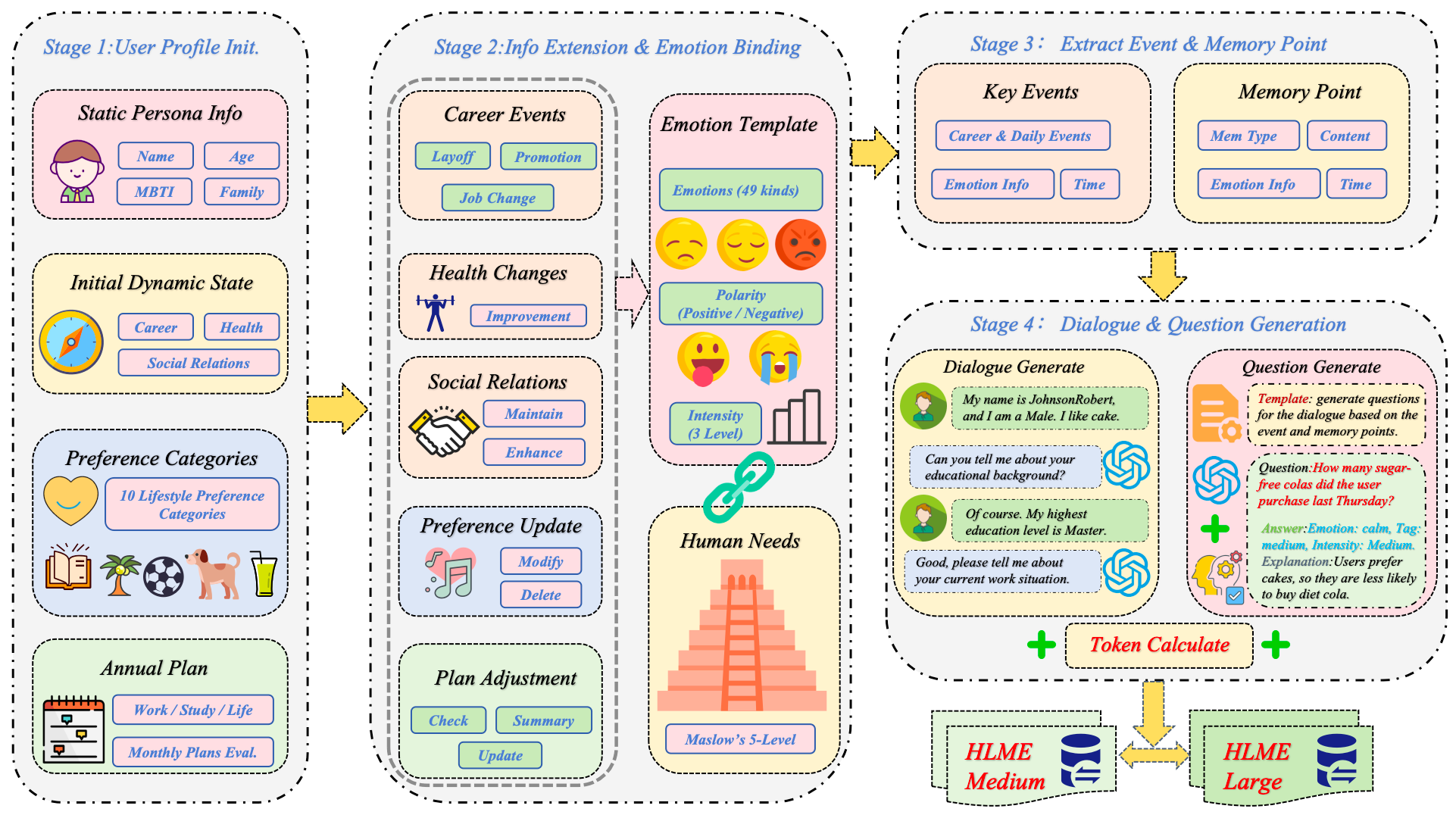}
    \caption{The HLME dataset construction process pipeline.}
    \label{fig:hlme_overall}
\end{figure*}

To evaluate the memory system’s ability to recognize, analyze, track, and reason about human emotions, we construct a high-quality human-like memory emotion evaluation dataset \textit{HLME}. To ensure both dataset quality and the ease of large-scale data construction, we design a four-stage dataset construction pipeline. The overall dataset generation process is illustrated in Figure~\ref{fig:hlme_overall}.

{\bf{Stage 1: Basic Information Generation}}
In the initial stage of user information construction, we leverage the persona dataset~\cite{ge2024scaling} as a seed source for sampling. For each randomly selected seed persona, we first establish the user’s static attributes, such as name, gender, age, educational background, and family composition. These basic attributes constitute the user’s initial state space and lay the foundation for subsequent generation of dynamic information.

We then construct the user’s dynamic attributes, which primarily include career status, health conditions, and social networks. The career status provides a detailed characterization of occupation, industry affiliation, job level or title, monthly income, and asset reserves; health conditions cover physical health, psychological state, history of chronic diseases, and underlying causes; and the social network describes the relational graph of social ties such as friends and colleagues.

In addition, based on fine-grained analysis of daily behaviors, we build ten categories of preference information, encompassing food, clothing, travel, and other aspects. Finally, integrating the above states, we formulate an annual plan that spans work, study, and daily life dimensions. This plan is further refined to a monthly granularity and equipped with concrete evaluation metrics to support dynamic adjustment and progress tracking in subsequent time steps. The comprehensive persona produced in this first stage serves as the cornerstone for driving emotion evolution and analysis.

{\bf{Stage 2: Extend Information and Emotion Binding}}
This stage aims to simulate the temporal evolution of user information and map it to emotional states. We first perform \emph{Information Extension}, which is carried out along three dimensions:
(1) \textit{Dynamic information evolution}: simulating changes in career trajectories, such as layoffs, job hopping, or promotions; fluctuations in health status, such as indicator abnormalities caused by lifestyle factors and subsequent recovery processes; and the evolution of social relationships, such as the maintenance, intensification, or dissolution of interpersonal bonds.
(2) \textit{Preference drift}: modeling preference revision or forgetting mechanisms induced by internal and external factors.
(3) \textit{Dynamic plan adjustment}: introducing a feedback mechanism that adjusts subsequent plans based on monthly execution outcomes, such as task impediments, and incorporating a quarterly review stage to dynamically update annual goals.

The second component is \emph{Emotion Binding}. We deeply integrate the extended information with an emotion model. Specifically, we adopt the Emotion Annotation and Representation Language (EARL) proposed by the Human–Machine Interaction Network on Emotion (HUMAINE)~\cite{schroder2005toward}, and construct templates covering 49 emotion categories along with their polarities, while defining three levels of emotional intensity: high, medium, and low. In addition, inspired by Maslow’s hierarchy of needs, which encompasses physiological, safety, social, esteem, and self-actualization needs, we establish mapping relationships between the user’s five fundamental needs, the extended events, and the base emotions. This multidimensional association mechanism provides prior conditions for generating dialogues with coherent logic and rich emotional depth.

{\bf{Stage 3: Extract Information Point and Generation Event}}
After completing emotion binding, this stage focuses on extracting core elements from complex contexts to guide dialogue generation.
We design a \emph{Memory Points} extraction mechanism that distills key metadata from the user persona, workflows, and derived information. Each memory point consists of the information type, specific content, the corresponding emotion label, and a timestamp.
In parallel, we identify and extract \emph{Key Events}, namely events that exert a significant impact on the user’s emotions, such as promotions with salary increases or sudden illnesses. These events serve as the core variables driving emotional state transitions and constitute critical evidence for evaluating a memory system’s capabilities in emotion understanding, attribution, and tracking. The combination of memory points and key events forms the structured input for generating high-quality dialogues.

{\bf{Stage 4: Generate Dialogue and Question}}
During the dialogue generation stage, we construct multi-turn dialogue data based on the previously generated event sequences. Each dialogue turn includes role identifiers for the User and Assistant, the corresponding textual content, and fine-grained emotion annotations that specify the type, label, and polarity of the emotional state. Furthermore, each entry is marked with timestamps and a unique dialogue ID. These data are packaged into a standardized test set, aiming to comprehensively evaluate the memory system’s emotion recognition and long-range memory capabilities.

In the question generation stage, to evaluate system performance from multiple perspectives, we design five categories of evaluation questions: simple factual questions, multi-hop complex reasoning questions, dynamic update questions, emotion conflict detection questions, and temporal reasoning questions. These questions are intended to assess the system’s core abilities in information extraction, memory updating, complex attribution, and conflict resolution.

In addition, we introduce a token accounting mechanism to measure the computational cost incurred by LLM when generating the complete dataset, and to monitor whether the context length exceeds the model’s window. Finally, we produce two versions of the dataset: a \emph{Medium} version and an \emph{Large} version. The Large version introduces noise information and extends dialogue length to construct challenging scenarios that exceed conventional context windows, thereby testing the system’s robustness in noisy environments and its ability to maintain memory over ultra-long contexts.

\section{Evaluation Framework of HLME}

We propose a LLM-based memory system evaluation task grounded in human emotion categories. Unlike traditional emotion recognition tasks, which predict an emotion label $y$ solely based on the input text $x$, this task requires the memory system to analyze, understand, store, reason about, and track the user’s emotional dynamics by leveraging information such as the user’s basic profile, dynamic information (e.g., changes in occupation, social relationships, health status, and family relations), preference information (e.g., dietary and clothing preferences), and annual plans. The goal of this task is to analyze the capabilities of several mainstream LLM-based memory systems (e.g., MemOS and zep) in emotion understanding, analysis, and memory, thereby providing guidance for the continual improvement and optimization of such memory systems.

\begin{figure*}[htbp]
    \centering
    \includegraphics[width=\linewidth]{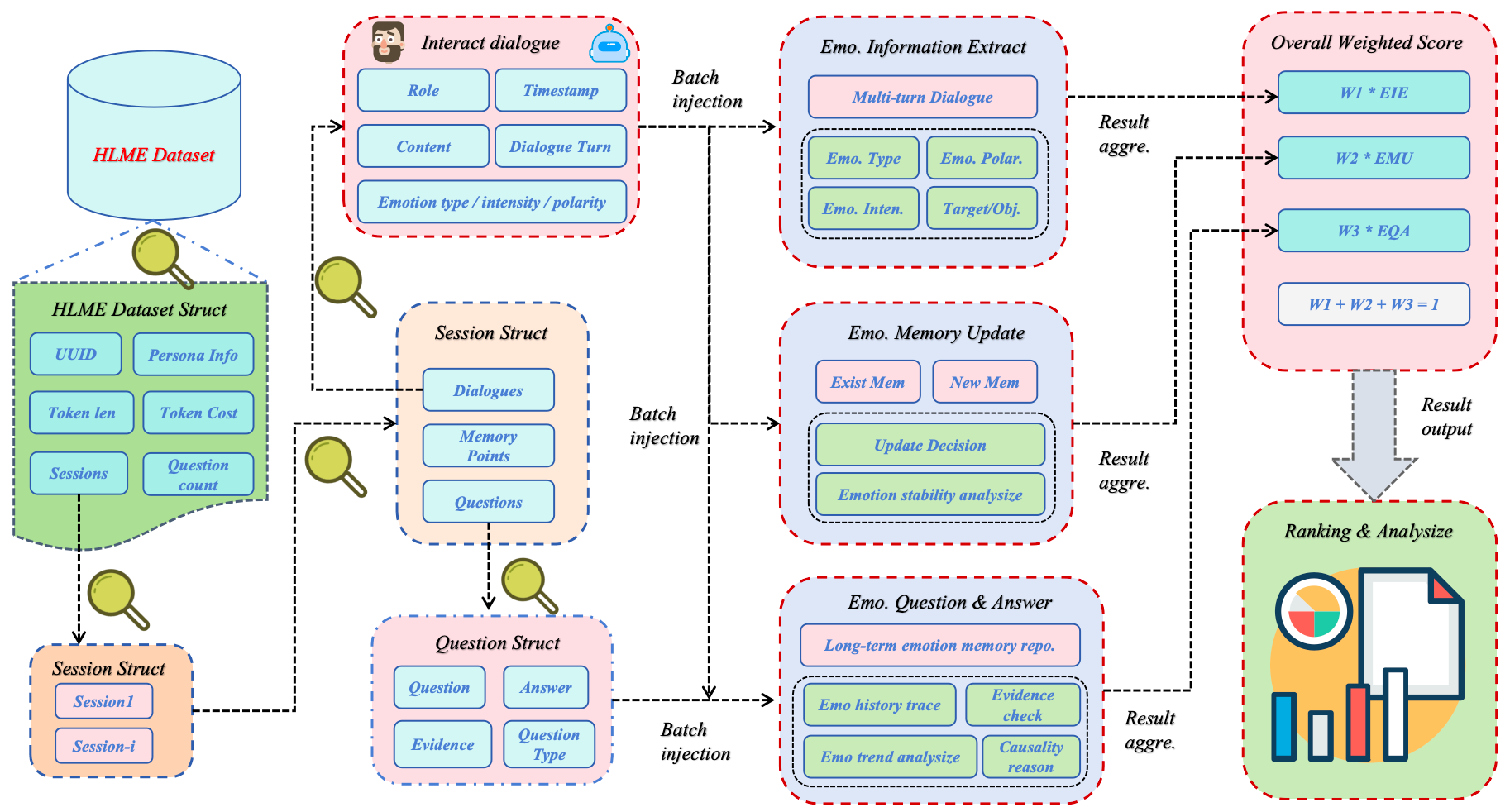}
    \caption{HLME dataset evaluation scheme.}
    \label{fig:evaluation_framework}
\end{figure*}

We categorize the evaluation tasks into three types: Emotion Information Extraction (EIE), Emotion Memory Update (EMU), and Emotion Question \& Answer (EQA). Detailed task descriptions and evaluation designs are presented in the subsequent subsections of this section. The complete evaluation framework is illustrated in Figure~\ref{fig:evaluation_framework}.

\subsection{Emotion Information Extract EIE}

\textbf{Task Description}: Given a multi-turn dialogue ($D$), the memory system is required to extract emotion-related facts that can be written into the memory system, including emotion type, polarity, intensity, and the target entity.

(1) \textbf{Emotion Classification Accuracy}: This metric measures the overall accuracy of the memory system in recognizing both explicit and implicit emotion categories.

\begin{equation}
\text{Acc} = \frac{1}{N} \sum_{i=1}^{N} \mathbbm{1}(y_i = \hat{y}_i)
\end{equation}

where $y_i$ denotes the ground-truth emotion label, $\hat{y}_i$ denotes the label predicted by the system, and $\mathbbm{1}(\cdot)$ is the indicator function.

(2) \textbf{Emotion Intensity MAE}: This metric measures the estimation error of emotion intensity by the memory system, directly reflecting its fine-grained emotion perception capability.

\begin{equation}
\text{MAE} = \frac{1}{N}\sum_{i=1}^{N} |\hat{s}_i - s_i|
\label{eq:f1-attr}
\end{equation}

where $s_i$ and $\hat{s}_i$ denote the ground-truth and predicted emotion intensity values, respectively (e.g., on a 1--3 scale).

(3) \textbf{Emotion Slot Extraction F1}: This metric evaluates whether the memory system can correctly extract complete emotion memory units (e.g., type + target + polarity), which directly determines the quality of memory writing.

\begin{equation}
\text{F1}_{\text{slot}} =
\frac{2 \cdot \left| \text{Slot}_{\text{pred}} \cap \text{Slot}_{\text{gold}} \right|}
{\left| \text{Slot}_{\text{pred}} \right| + \left| \text{Slot}_{\text{gold}} \right|}
\end{equation}

where $\text{Slot}$ denotes a set composed of multiple attributes.

\subsection{Emotion Memory Update EMU}
\textbf{Task Description}: Given the existing memory ($M_t$) and a new dialogue ($D_{t+1}$), the memory system is required to determine whether the emotional state should be updated and to generate the updated memory accordingly.

(1) \textbf{Update Decision Accuracy}: This metric measures whether the memory system can correctly determine whether the emotional information associated with a given dialogue topic should be updated, reflecting its sensitivity to emotional changes.

\begin{equation}
\text{Acc}_{\text{update}} = \frac{1}{N} \sum_{i=1}^{N} \mathbbm{1}(\hat{o}_i = o_i)
\end{equation}

where $o_i \in \{0, 1\}$ denotes the ground-truth label of the $i$-th memory, and $\hat{o}_i$ denotes the label predicted by the memory system.

(2) \textbf{Intensity Delta MAE}: This metric evaluates whether the memory system correctly captures the magnitude of emotional change, rather than only the current state, thereby reflecting its capability for dynamic emotion modeling.

\begin{equation}
\text{MAE}_{\Delta} =
\frac{1}{N} \sum_{i=1}^{N}
\left|
(\hat{s}_{t+1} - \hat{s}_t) - (s_{t+1} - s_t)
\right|
\end{equation}

where $s_t$ denotes the ground-truth emotion intensity at time $t$, while $\hat{s}_t$ denotes the intensity predicted by the system at time $t$.

(3) \textbf{Memory Stability Score (MSS)}: This metric measures the system’s ability to preserve previously correct memories without erroneous overwriting when exposed to interfering information that does not involve emotional changes.

\begin{equation}
\text{MSS} = 1 - \frac{|\text{Mem}_{\text{err\_upd}}|}{|\text{Mem}_{\text{static}}|}
\end{equation}

where $\text{Mem}_{\text{err\_upd}}$ denotes the number of erroneously updated memories, and $\text{Mem}_{\text{static}}$ denotes the total number of emotion facts that should remain unchanged. This metric reflects the system’s ability to prevent memory drift.

\subsection{Emotion Question \& Answer EQA}
\textbf{Task Description}: The memory system is required to answer questions about the user’s historical emotional states, their temporal evolution trends, and underlying causes based on the memory repository.

(1) \textbf{Emotion QA Accuracy}: This metric evaluates whether the memory system can provide factually correct answers to emotion-related questions.

\begin{equation}
\text{Acc}_{\text{QA}} = \frac{1}{N} \sum_{i=1}^{N} \mathbbm{1}(\hat{a}_i = a_i)
\end{equation}

where $a_i$ denotes the reference answer to questions about emotional history or trends, and $\hat{a}_i$ denotes the answer generated by the memory system.

(2) \textbf{Evidence Grounding F1}: This metric evaluates whether the memory system’s answers are grounded in the correct emotional memory evidence, thereby preventing hallucinated responses.

\begin{equation}
\text{F1}_{\text{evidence}} =
\frac{2 \cdot P_e \cdot R_e}{P_e + R_e}
\end{equation}

where $P_e$ and $R_e$ denote the precision and recall of the retrieved evidence snippets, respectively.

\subsection{Overall Weighted Score}
By computing a weighted sum of the scores from the three components above, we obtain an overall score, which reflects the differences among systems through comparative performance.

\begin{equation}
\text{Score}_{\text{overall}} =
\sum_{k=1}^{3} \alpha_k \cdot \text{Score}_k,
\quad \text{s.t. } \sum_{k=1}^{3} \alpha_k = 1
\end{equation}

where $\alpha_k$ denotes the weight assigned to each task, with $\sum \alpha_k = 1$, and $\overline{\text{Score}}_k$ denotes the normalized mean of the metrics within each task.

\section{Experiments}

\subsection{Experiments Setup}

We constructed a humanoid memory–emotion evaluation dataset to assess the ability of memory systems to process, track, and reason about emotional information during long-term and short-term memory handling. We designed three major evaluation tasks to examine memory systems’ capabilities in emotional information extraction, updating, and question answering. We evaluated six systems, including MemOS \cite{li2025memos_long}, MemoBase \cite{memobase2025}, Mem0 \cite{chhikara2025mem0}, Mirix \cite{wang2025mirix}, and Letta \cite{packer2023memgpt}.

We adopted the \textit{LLM-as-a-Judge} evaluation paradigm, using GPT-4o-mini as the evaluation model. We assessed the six aforementioned memory systems using two versions of the dataset, namely HLME-Medium and HLME-Long, together with a uniformly designed evaluation template. The evaluation focused on the systems’ capabilities in emotional information extraction, updating, and question answering. In addition, we analyzed the retrieval performance of the memory systems by varying the top-(k) settings.

\subsection{The Experimental Results Analysis}

This section evaluates five representative memory systems within the HLME framework through a multi-faceted analysis. We first benchmark their performance on three primary tasks, namely emotional information extraction (EIE), emotional memory updating (EMU), and emotional question answering (EQA), across two dataset versions to distinguish their capabilities in static emotion perception and dynamic emotional tracking. To assess practical viability, we further analyze computational efficiency by measuring the overhead associated with memory insertion and memory retrieval. Finally, we examine the impact of the Top-(K) retrieval window on EQA performance, illustrating how variations in contextual density influence the systems’ ability to leverage evidence for emotion-oriented question answering.

\subsubsection{Overall Evaluation of HLME}
The experimental results of different memory systems are summarized in Table~\ref{tab:hlme_overall}.

\begin{table*}[htbp]
\caption{Overall emotion-related performance of different memory systems on Medium (Med.) and Long.
$\uparrow$ / $\downarrow$ indicate higher / lower is better.
Best results are in \textbf{bold}, worst results are \underline{underlined}.}
\label{tab:hlme_overall}
\centering
\setlength{\tabcolsep}{3.5pt}
\resizebox{\textwidth}{!}{
\begin{tabular}{llcccccccccc}
\toprule
\multirow{2}{*}{\textbf{Model}} &
\multirow{2}{*}{\textbf{Setting}} &
\multicolumn{3}{c}{\textbf{EIE}} &
\multicolumn{3}{c}{\textbf{EMU}} &
\multicolumn{2}{c}{\textbf{EQA}} &
\textbf{Overall} \\
\cmidrule(lr){3-5}
\cmidrule(lr){6-8}
\cmidrule(lr){9-10}
\cmidrule(lr){11-11}
&
& Acc$\uparrow$ & MAE$\downarrow$ & F1$_{\text{slot}}\uparrow$
& Acc$_{\text{update}}\uparrow$ & MAE$_\Delta\downarrow$ & MSS$\uparrow$
& Acc$_{\text{QA}}\uparrow$ & F1$_{\text{evidence}}\uparrow$
& Score$\uparrow$ \\
\midrule
Mem0 & Med.
& \underline{0.4534} & \underline{1.6667} & \underline{0.0871}
& \underline{0.0011} & \underline{1.9977} & 0.7762
& 0.50 & 0.6082
& \underline{0.3574} \\
Letta & Med.
& 0.8082 & 0.9942 & 0.3632
& \textbf{0.3949} & \textbf{1.21} & 0.9543
& \textbf{0.6265} & 0.6945
& 0.5547 \\
MemOS & Med.
& 0.707 & 0.8837 & 0.2049
& 0.0205 & 1.9589 & 0.7132
& 0.5531 & \textbf{0.7074}
& 0.4378 \\
Mirix & Med.
& \textbf{0.9629} & \textbf{0.1802} & \textbf{0.8774}
& 0.3378 & 1.3242 & \textbf{0.9794}
& \underline{0.4352} & \underline{0.5371}
& \textbf{0.5952} \\
MemoBase & Med.
& 0.8872 & 0.386 & 0.4229
& 0.0433 & 1.9132 & \underline{0.3687}
& 0.5271 & 0.662
& 0.4541 \\
\midrule
\addlinespace[2pt]
Mem0 & Large
& \underline{0.5695} & \underline{1.1913} & \underline{0.1443}
& \underline{0.0068} & 1.9863 & 0.7648
& \underline{0.4698} & 0.5875
& \underline{0.3836} \\
Letta & Large.
& \textbf{0.8448} & 0.8103 & 0.4492
& \textbf{0.173} & \textbf{1.6538} & 0.9615
& 0.4848 & 0.6515
& \textbf{0.5122} \\
MemOS & Large
& 0.6024 & 1.0144 & 0.202
& 0.0228 & \underline{1.9543} & 0.5479
& \textbf{0.5737} & \textbf{0.7039}
& 0.4089 \\
Mirix & Large
& 0.7028 & 0.9127 & \textbf{0.5129}
& 0.049 & 1.9018 & \textbf{0.9965}
& 0.4774 & \underline{0.505}
& 0.4633 \\
MemoBase & Large
& 0.6798 & \textbf{0.8082} & 0.3076
& 0.0251 & 1.9497 & \underline{0.3584}
& 0.4743 & 0.6261
& 0.3919 \\
\bottomrule
\end{tabular}
}
\end{table*}

The results show substantial performance disparities across systems under different context lengths. Mirix (Medium) achieves the highest overall score in medium-length conversations, demonstrating its strong accuracy in handling short- to mid-range interactions. In contrast, in long-context scenarios, Letta (Large) ranks first in terms of overall performance, indicating that its operating-system-inspired memory management mechanism is robust to emotional evolution over extended context windows. By comparison, Mem0 consistently ranks last under both the Medium and Large dataset settings, suggesting that its simple vector-retrieval-based architecture struggles to cope with complex emotional tracking tasks.

The Emotional Information Extraction (EIE) task is designed to evaluate the accuracy of memory systems in emotion perception. On the Medium version of the dataset, Mirix achieves an emotion extraction accuracy (Acc) exceeding 90\%, with an F1 score for emotional memory unit extraction above 80\%, significantly outperforming other systems. This result indicates that its agent collaboration mechanism is particularly effective at capturing explicit emotional facts. Letta attains the highest extraction accuracy on the Large dataset, reflecting its robustness to noise in long-context environments. In contrast, Mem0 performs the worst on this task, highlighting that systems lacking structured graphs or explicit memory management are prone to losing fine-grained emotional attributes.

The Emotional Memory Updating (EMU) task evaluates the sensitivity and stability of memory systems in tracking dynamic emotional changes. Letta shows a clear advantage in update decision accuracy (Acc$_{\text{update}}$), especially on the Large dataset, where its performance far exceeds that of other systems. This advantage stems from its active write-permission mechanism, which enables the system to proactively assess and overwrite outdated core memories through an internal “inner monologue,” analogous to an operating system. In contrast, Mem0 achieves an update accuracy close to zero, confirming that it effectively operates in a read-only or append-only mode and is unable to handle changes in emotional states. In terms of stability (MSS), both Mirix and Letta maintain exceptionally high scores above 95\%, indicating strong resistance to interference that could corrupt correct memories. MemoBase, however, exhibits very low stability scores, suggesting that its overly aggressive compression strategy leads to frequent erroneous updates or hallucinations; in prioritizing coverage, MemoBase sacrifices memory accuracy.

The Emotional Question Answering (EQA) task assesses memory systems’ capabilities in emotional reasoning and provenance tracing. MemOS performs best on this task, particularly on the Large dataset, where it achieves the highest accuracy and evidence-tracing F1 score. This result strongly validates the effectiveness of MemOS’s \textit{MemCube} hierarchical scheduling mechanism, which allows the system to ground its answers in verifiable memory evidence and effectively mitigate hallucination issues common in generative models. Notably, although Mirix excels during the extraction stage, its performance in the EQA task is relatively modest, indicating potential information loss when transforming stored structured representations into reasoning-based answers.

Overall, the HLME evaluation framework clearly delineates the capability boundaries of existing memory systems. Letta excels at long-term dynamic updating, MemOS specializes in precise retrieval and provenance tracing, and Mirix demonstrates strong performance in short- to mid-term static extraction. At present, no single system achieves comprehensive superiority across all dimensions. This observation underscores a central open challenge in memory system design: how to strike an effective balance between high-sensitivity perception and highly stable long- and short-term memory interaction.

\subsubsection{Efficiency of Execution}
Among the three evaluation tasks introduced in Section~5, emotional information extraction and emotional memory updating are write-intensive operations. These tasks require memory systems to extract factual information from conversations and rely on a memory controller to determine whether existing memories should be overwritten. In contrast, emotional memory question answering is a read-intensive operation, in which the memory system must identify relevant evidence from a large memory repository in response to a query and return the most relevant top-(k) memory entries.

The execution efficiency of the aforementioned systems is reported in Table~\ref{tab:retrieval_time}.

\begin{table}[htbp]
\centering
\setlength{\tabcolsep}{6pt}
\caption{Memory operation latency on Medium and Large version datasets, The time unit in the experimental results is expressed in seconds.\textbf{Add} represents the time when all the data is added to the memory system, while \textbf{search} represents the total time for multiple searches such as top-10, top-20, and top-50. Best results are in \textbf{bold}, worst results are \underline{underlined}.}
\label{tab:retrieval_time}
\begin{tabular}{llcc}
\toprule
\textbf{Dataset} & \textbf{System} & \textbf{Add} & \textbf{Search} \\
\midrule
\multirow{5}{*}{\textbf{Medium}}
& Mem0     
& 141.10 
& \textbf{1624.33} \\
& Letta    
& 1085.86 
& 4075.66 \\
& MemOS    
& 622.88 
& 1665.64 \\
& Mirix    
& \textbf{123.83} 
& 4821.08 \\
& MemoBase 
& \underline{3478.41} 
& \underline{4895.94} \\
\midrule
\multirow{5}{*}{\textbf{Large}}
& Mem0     
& 356.01 
& 2868.70 \\
& Letta    
& \underline{3905.85} 
& \underline{4732.06} \\
& MemOS    
& 1094.29 
& \textbf{1777.35} \\
& Mirix    
& \textbf{230.28} 
& 3697.04 \\
& MemoBase 
& 1562.38 
& 3895.41 \\
\bottomrule
\end{tabular}
\end{table}

During the memory writing phase, the primary sources of time overhead arise from factual extraction and index construction. Mirix demonstrates the highest write efficiency, benefiting from the parallel processing capability of its multi-agent architecture, which enables rapid distribution and processing of fragmented information. Mem0 follows closely, as its straightforward vector-append strategy avoids complex logical decision-making and thus maintains low latency. By contrast, Letta incurs the highest write-time cost on the Large dataset. This overhead is an inherent consequence of its system introspection mechanism, in which the system repeatedly reasons about whether newly acquired information conflicts with existing core memories. Such autonomous decision-making requires substantial inference time. Notably, MemoBase exhibits abnormally high latency on the Medium dataset, indicating that its streaming compression and encoding mechanisms introduce significant computational bottlenecks when handling dense short- to mid-length contexts. As a result, its write efficiency is even lower than that of the more logically complex Letta.

In the emotional memory question answering task, time consumption is primarily driven by the expansion of the semantic retrieval scope and the aggregation of evidential information for answering. MemOS exhibits outstanding scalability at this stage. Although it is slightly slower than Mem0 on the Medium dataset, MemOS achieves the lowest retrieval latency on the larger Large dataset. This result strongly demonstrates the effectiveness of its hierarchical storage mechanism, which employs L1/L2 caching. Frequently accessed “hot” memories are retained in fast cache layers, while only infrequently accessed “cold” memories trigger full-database scans, enabling rapid response even at scale. In contrast, both MemoBase and Letta incur consistently high retrieval latency, indicating that complex context reconstruction or recursive retrieval introduces substantial I/O overhead during the read phase. The retrieval performance of Mirix is also constrained by its architectural design. On the Large dataset, within long-context settings, communication overhead among multiple agents becomes a critical bottleneck, as repeated inter-agent coordination to determine memory ownership leads to exponentially increasing retrieval latency.

\subsubsection{Retrieval performance }
In the preceding evaluations, we uniformly adopted a Top-10 retrieval window. To further investigate the impact of retrieval scope on complex emotional reasoning, we expanded the context window to \textit{Top-20} and \textit{Top-50}, respectively. The detailed results are presented in Table~\ref{tab:dataset_overall}.

\begin{table}[htbp]
\caption{Emotion QA (EQA) performance of different memory systems on Medium and Large settings.\textbf{T20} and \textbf{T50} represent top-20 and top-50.Best results are in \textbf{bold}, worst results are \underline{underlined}.}
\label{tab:dataset_overall}
\centering
\setlength{\tabcolsep}{6pt}
\renewcommand{\arraystretch}{1.15}
\resizebox{\columnwidth}{!}{
\begin{tabular}{lccccc}
\toprule
\multirow{2}{*}{\textbf{Dataset}} 
& \multirow{2}{*}{\textbf{Model}} 
& \multicolumn{2}{c}{\textbf{EQA (T20)}} 
& \multicolumn{2}{c}{\textbf{EQA (T50)}} \\
\cmidrule(lr){3-4}
\cmidrule(lr){5-6}
& 
& Acc$_{\text{QA}}$ 
& F1$_{\text{evid.}}$ 
& Acc$_{\text{QA}}$ 
& F1$_{\text{evid.}}$ \\
\midrule

\multirow{5}{*}{Medium}
& Mem0      & 0.505 & 0.606 & 0.486 & 0.597 \\
& Letta     & \textbf{0.620} & \underline{0.439} & \textbf{0.646} & 0.716 \\
& MemOS     & 0.532 & \textbf{0.690} & 0.563 & \textbf{0.741} \\
& Mirix     & \underline{0.462} & 0.562 & \underline{0.474} & \underline{0.570} \\
& MemoBase  & 0.548 & 0.689 & 0.568 & 0.694 \\
\midrule

\addlinespace[2pt]
\multirow{5}{*}{Large}
& Mem0      & 0.476 & 0.599 & 0.521 & 0.632 \\
& Letta     & \underline{0.455} & 0.591 & \underline{0.482} & 0.630 \\
& MemOS     & \textbf{0.565} & \textbf{0.707} & \textbf{0.577} & \textbf{0.713} \\
& Mirix     & 0.500 & \underline{0.512} & 0.494 & \underline{0.520} \\
& MemoBase  & 0.532 & 0.652 & 0.536 & 0.673 \\

\bottomrule
\end{tabular}
}
\end{table}

The experimental results indicate that, as the retrieval window is expanded from Top-20 to Top-50, the majority of memory systems exhibit a positive increase in evidence-tracing performance for question answering (F1$_{\text{evid.}}$). Taking MemOS as an example, its F1 score on the Medium dataset improves from 0.6904 to 0.7408. This trend suggests that a broader retrieval scope can capture more scattered emotional cues, thereby providing memory systems with richer evidential support.

However, continuously enlarging the retrieval window does not lead to unbounded performance gains. For instance, Letta shows a noticeable decline in accuracy on the Large dataset when the retrieval window becomes excessively large. This behavior indicates that its memory management mechanism is susceptible to attention interference from irrelevant information when confronted with overly long historical contexts, resulting in disruptions within the reasoning chain.

Overall, the experimental findings clearly demonstrate that high-performing memory systems, such as MemOS, must not only be capable of retrieving relevant information but also possess strong robustness against interference to retrieve precise evidence. The ability to effectively handle user memories across varying temporal spans represents a core aspect of the value of memory systems.

\section{Conclusion}
Memory systems address the limitation of large language models in tracking and updating long-term historical memory. However, substantial research opportunities remain in how memory systems handle emotion-related interactions. We introduce a humanoid memory–emotion evaluation dataset designed to assess the capabilities of current mainstream memory systems in emotional extraction, emotional updating, and emotion-oriented question answering. Through our dataset and benchmark, we provide a new direction for advancing emotional processing in memory systems. The ability of memory systems to perceive and process emotions constitutes a crucial component in realizing AI systems with human-like warmth.

\section*{Limitations and Future Work}
This study evaluates a limited set of memory system architectures. While these systems reflect current state-of-the-art designs, broader cross-architectural evaluations are necessary to fully assess the robustness and generalizability of the benchmark. In addition, many memory systems lack native support for conversational memory APIs, which complicates comprehensive evaluation. 

Future work will focus on improving dataset quality and benchmark generality through partial manual annotation. We also plan to incorporate additional backbone models to examine how different base models influence evaluation outcomes. Moreover, we will expand task coverage and refine evaluation metrics to enable a more detailed, objective, and comprehensive assessment of emotional processing capabilities in memory systems.

\bibliography{reference}

\appendix



\end{document}